\documentclass[sigconf]{acmart}
\usepackage{subcaption}
\usepackage{listings}
\usepackage{algorithm}
\usepackage{algpseudocode}

\AtBeginDocument{%
  \providecommand\BibTeX{{%
    \normalfont B\kern-0.5em{\scshape i\kern-0.25em b}\kern-0.8em\TeX}}}

\copyrightyear{2021}
\acmYear{2021}
\setcopyright{acmcopyright}\acmConference[FPGA '21]{Proceedings of the 2021 ACM/SIGDA International Symposium on Field Programmable Gate Arrays}{February 28-March 2, 2021}{Virtual Event, USA}
\acmBooktitle{Proceedings of the 2021 ACM/SIGDA International Symposium on Field Programmable Gate Arrays (FPGA '21), February 28-March 2, 2021, Virtual Event, USA}
\acmPrice{15.00}
\acmDOI{10.1145/3431920.3439296}
\acmISBN{978-1-4503-8218-2/21/02}

\newcommand{\Name}{FracBNN}

\begin{document}
\fancyhead{}
\title{{\Name}: Accurate and FPGA-Efficient Binary Neural Networks with Fractional Activations}



\author{
Yichi Zhang$^{1}$,
Junhao Pan$^{2}$,
Xinheng Liu$^{2}$,
Hongzheng Chen$^{3,1}$,
Deming Chen$^{2}$,
Zhiru Zhang$^{1}$
}

\affiliation{
\LARGE{%
	\institution{
    $^1$Cornell University\quad 
	$^2$University of Illinois Urbana-Champaign\quad
	$^3$Sun Yat-sen University
    }}
}
\email{{yz2499, zhiruz}@cornell.edu}

\keywords{Deep Learning; Binary Neural Networks; FPGA Acceleration; HLS 
}

\renewcommand{\shortauthors}{Yichi Zhang, et al.}
\renewcommand{\authors}{Yichi Zhang, Junhao Pan, Xinheng Liu, Hongzheng CHen, Deming Chen, Zhiru Zhang 
}

\begin{abstract}
  Binary neural networks (BNNs) have 1-bit weights and activations. 
  Such networks are well suited for FPGAs, as their dominant computations are bitwise arithmetic and the memory requirement is also significantly reduced. 
  However, compared to start-of-the-art compact convolutional neural network (CNN) models, BNNs tend to produce a much lower accuracy on realistic datasets such as ImageNet. In addition, the input layer of BNNs has gradually become a major compute bottleneck, because it is conventionally excluded from binarization to avoid a large accuracy loss. 
  
  This work proposes {\Name}, which exploits fractional activations to substantially improve the accuracy of BNNs. Specifically, our approach employs a dual-precision activation scheme to compute features with up to two bits,  using an additional sparse binary convolution. We further binarize the input layer using a novel thermometer encoding. Overall, {\Name} preserves the key benefits of conventional BNNs, where all convolutional layers are computed in pure binary MAC operations (BMACs). 
  We design an efficient FPGA-based accelerator for our novel BNN model that supports the fractional activations. To evaluate the performance of {\Name} under a resource-constrained scenario, we implement the entire optimized network architecture on an embedded FPGA (Xilinx Ultra96 v2). Our experiments on ImageNet show that {\Name} achieves an accuracy comparable to MobileNetV2, surpassing the best-known BNN design on FPGAs with an increase of 28.9\% 
  in top-1 accuracy and a 2.5$\times$ reduction in model size. {\Name} also outperforms a recently introduced BNN model with an increase of 2.4\% in top-1 accuracy while using the same model size.
  On the embedded FPGA device, {\Name} demonstrates the ability of real-time image classification. 
\end{abstract}

\maketitle

\section{Introduction}



\begin{figure}
  \centering
  \includegraphics[width=\linewidth]{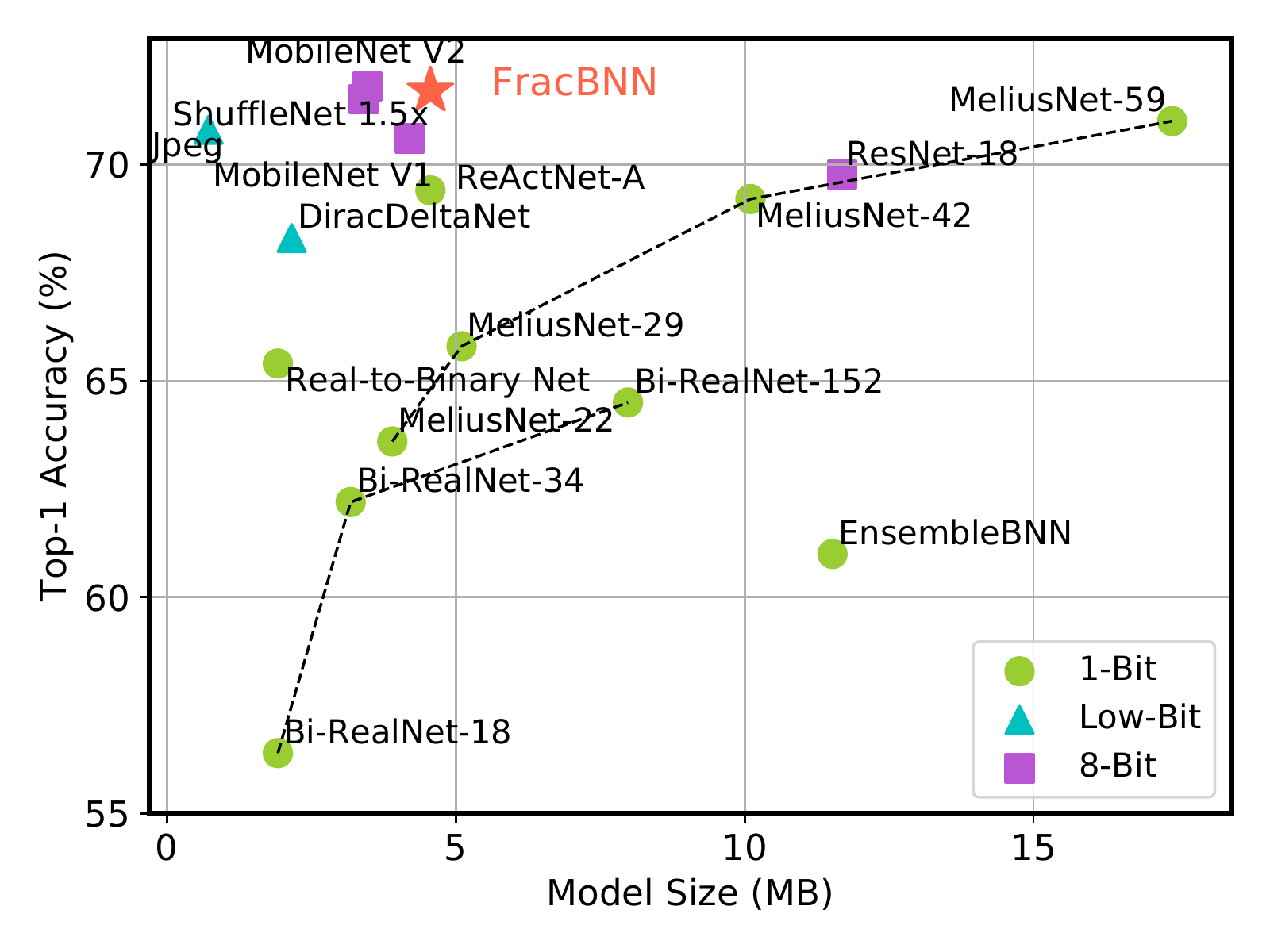}
  \vspace{-15pt}
  \caption{Comparison of ImageNet top-1 accuracy and model size of various compact CNN models.}
  \label{fig:benchmarks}
  \vspace{-15pt}
\end{figure}

Binary neural network (BNN) is a promising approach to improving the efficiency of deep learning execution, especially for CNNs \cite{qin2020survey, hubara2016bnn, rastegari2016xnornet, Liu_2018_ECCV}. With binarized weights and activations, the dominant computations of a BNN model are binary multiply-accumulate (BMAC) operations, which can be implemented in a highly hardware-efficient way using XNORs and population counts (popcnt). The extreme quantization can also reduce the memory requirement for storing the model.

FPGAs are a great match for implementing the BNN models, as the BMAC operations can be mapped and executed on the LUT-based logic fabric in a massively parallel fashion. The reduced memory footprint is also attractive since FPGAs tend to have limited on-chip SRAM capacity. 
For these reasons, extensive studies have been devoted to the FPGA acceleration of BNNs.
Umuroglu et al.~\cite{yaman2017finn} and Zhao et al.~\cite{zhao2017bnnacc} are among the first to implement an off-the-shelf binarized VGG network on CIFAR-10~\cite{hubara2016bnn} using high-level synthesis (HLS).
Later, several hardware-friendly BNN models are proposed to make inference more efficient~\cite{guo2018fbna,li2018pcbnn,ghasemzadeh2018rebnet,liang2018fpbnn,qin2020survey}. For example, \cite{guo2018fbna} improves the latency of the BNN CIFAR-10 accelerator to 1.9ms per inference on a Xilinx Zynq device; FP-BNN~\cite{liang2018fpbnn} implements a binarized AlexNet~\cite{krizhevsky2012alexnet} for the ImageNet dataset and delivers a latency of 1.16ms on an Intel Stratix V FPGA.

While BNNs provide obvious benefits in hardware implementation, it is facing two major challenges that are detailed as follows. 

\textbf{BNNs produce a low accuracy on realistic datasets --} Hubara et al.~\cite{hubara2016bnn} pioneered the recent advances in BNNs. This work shows competitive accuracy on small datasets such as MNIST and CIFAR-10. Unfortunately, a binarized AlexNet on ImageNet only achieves a top-1 accuracy of 36.1\%, which is more than 20\% lower (in absolute difference) than the original full-precision model. 
The state-of-the-art ImageNet BNN accelerator on FPGAs is based on the same model~\cite{liang2018fpbnn}. Not surprisingly, its accuracy remains very low at 42.9\%. 
Most recently, ReActNet~\cite{liu2020reactnet} modifies the MobileNet V1 architecture~\cite{howard2017mobilenets} and dramatically increases the accuracy to 69.4\% through activation shifting and reshaping. However, this model has as many as 29.3 million parameters (29.3 million bits).
In contrast, compact CNN models such as MobileNet V2~\cite{Sandler_2018_CVPR} can achieve an accuracy of 72\% with 3.4 million parameters (27.2 million bits). 

\textbf{The first convolutional layer is not binarized --}
Existing BNN models commonly use floating-point weights and activations in the input layer to avoid a large degradation in accuracy ~\cite{rastegari2016xnornet, Liu_2018_ECCV, bethge2020meliusnet, liu2020reactnet, yang2019synetgy}. 
The first layer copes with three input channels, thus involving fewer floating-point MAC operations compared to other layers in a conventional CNN. 
On embedded FPGA devices, however, it is difficult to exploit high parallelism to compute the floating-point input layer due to limited DSP resources.
Moreover, a dedicated floating-point convolution engine must be instantiated to execute the input layer, which is not resource-efficient since this engine cannot be reused by other layers. 
Some prior efforts have attempted to quantize the input layer using fixed-point types~\cite{hubara2016bnn, zhao2017bnnacc}. Unfortunately, these techniques typically incur a nontrivial accuracy loss, especially on realistic datasets such as ImageNet.

To overcome the aforementioned challenges, we propose \emph{\Name}, an efficient and accurate binary neural network with fractional activations. All convolutional layers in {\Name} are computed in pure BMACs (input layer included).
We first construct a baseline BNN model motivated by ReActNet~\cite{liu2020reactnet}. 
To improve the accuracy, we compute an extra sparse binary convolutional layer to update a fraction of the features using two bits, thus exploiting fractional activations.
As shown in Figure~\ref{fig:benchmarks}, {\Name} outperforms state-of-the-art BNNs and low-bitwidth networks by a large margin. In particular, it achieves MobileNetV2-level accuracy with a competitive model size. 
We further design an efficient FPGA-based BNN accelerator that supports fractional activations. We implement the entire {\Name} accelerator using HLS, and accelerate the inference on an embedded FPGA (Xilinx Ultra96 v2). 
{\Name} demonstrates the ability of real-time image classification by achieving a frame rate of 48.1 fps.

Our main technical contributions are as follows:
\begin{itemize}
    \item 
    We propose {\Name}, an accurate and efficient BNN architecture with fractional activations, where all convolutional layers are computed in BMACs.
    On ImageNet, {\Name} outperforms the best-known FPGA-targeted BNN by 28.9\% and the state-of-the-art ReActNet model by 2.4\% in top-1 accuracy. For the first time, we show a CNN with pure BMACs can achieve the same level of accuracy with MobileNet V2.
    
    \item In an end-to-end trainable BNN, we propose to use thermometer encoding to preprocess the images and binarize the input layer. We show that thermometer encoding helps with preserving the feature similarity,
    thus incurring minimal accuracy degradation.
    
    \item We design a novel FPGA-based BNN accelerator that supports fractional activations. We implement our design in HLS and demonstrate real-time performance for inference on an embedded FPGA. 
    In terms of frame rate, our FPGA implementation outperforms the most accurate BNN accelerator for CIFAR-10~\cite{zhao2017bnnacc} 
    and a state-of-the-art 4-bit CNN accelerator for ImageNet~\cite{yang2019synetgy}. 
\end{itemize}

\section{BNN Preliminaries}
\label{sec:background}

In this section, we first describe conventional BNN models used for FPGA implementation. We then introduce a recently proposed BNN model that has achieved a dramatic improvement in accuracy.

\subsection{Conventional BNN Models}
\begin{figure}[ht]
  \centering
  \includegraphics[width=\linewidth]{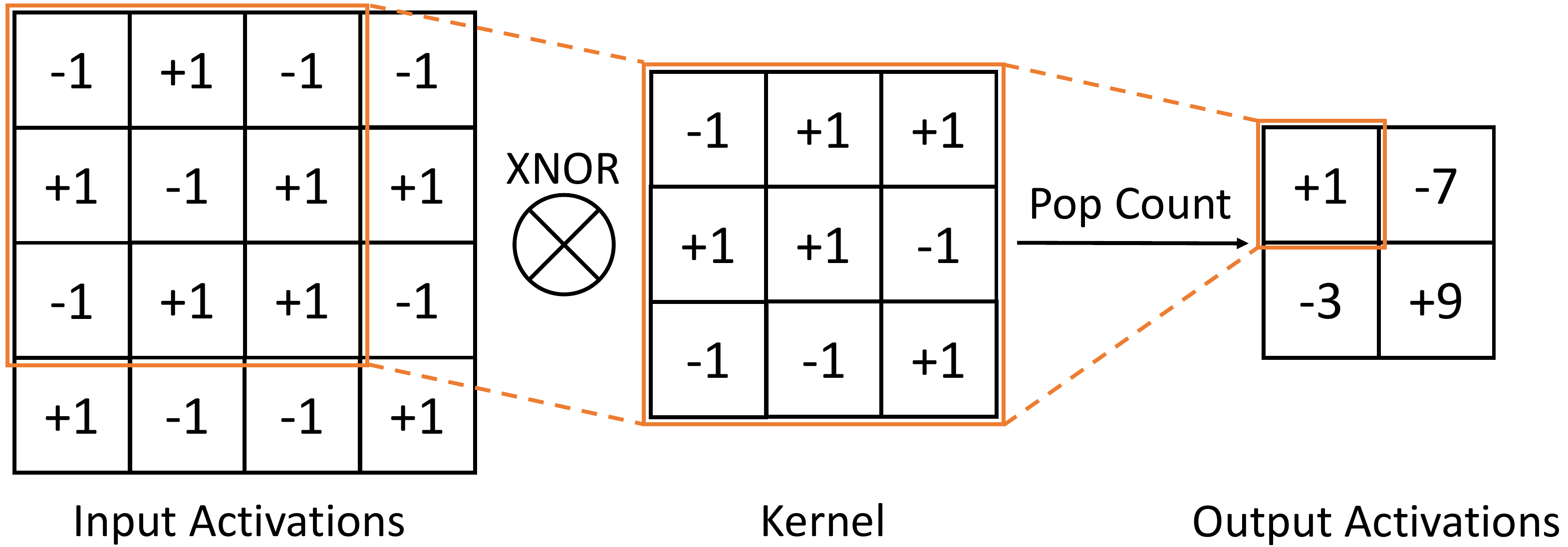}
  \caption{Convolution in a BNN model.}
  \label{fig:bnn-conv}
\end{figure}


In BNNs, we define a binary convolutional layer as 
\begin{displaymath}
    \textbf{O} = \textbf{W}^{b} * \textbf{X}^{b}
\end{displaymath}
$\textbf{O}$ is the convolution results. $\textbf{W}^{b}$ is the kernel weight tensor, and $\textbf{X}^{b}$ is the activation tensor. Both $\textbf{W}^{b}$ and $\textbf{X}^{b}$ are binarized to either $-1$ or $+1$ using the sign function. Specifically, 
\begin{displaymath}
    w^b = \text{sign}(w^r) =
    \left\{\begin{matrix}
    +1 & w^r \ge 0 \\ 
    -1 & w^r  <  0
    \end{matrix}\right.
    , \quad
    x^b = \text{sign}(x^r) =
    \left\{\begin{matrix}
    +1 & x^r \ge 0 \\ 
    -1 & x^r  <  0
    \end{matrix}\right.
\end{displaymath}
The superscripts $b$ and $r$ denote binary and real values, respectively.

As shown in Figure~\ref{fig:bnn-conv}, due to the binarization of weights and activations, a multiplication and addition (MAC) can be implemented as a bitwise XNOR  followed by a popcnt. We can therefore rewrite the binary convolutional layer as
\begin{displaymath}
    \textbf{O} = \text{popcnt} ( \text{XNOR} (\textbf{W}^{b}, \textbf{X}^{b} ) )
\end{displaymath}
Since XNOR and popcnt operations can easily be mapped and parallelized on the LUT fabric, it is highly efficient to perform the BNN inference on FPGAs.



Note that the BNN models for realistic datasets such as ImageNet implemented by existing FPGA accelerators~\cite{liang2018fpbnn, zhao2017bnnacc} are typically binarized AlexNet~\cite{krizhevsky2012alexnet} or VGG~\cite{simonyan2015vgg}. The stacked building block consists of a sign function, a binary convolutional layer, and a normalization layer in sequence. 
Although the binarized layers are efficient on FPGAs, the accuracy of the aforementioned binarized models is low.
Moreover, the entire model size is usually more than 10 MB, which exceeds the typical on-chip SRAM capacity of modern embedded FPGAs. 

\subsection{An Improved BNN Model}

\begin{figure}[h]
  \centering
  \includegraphics[width=\linewidth]{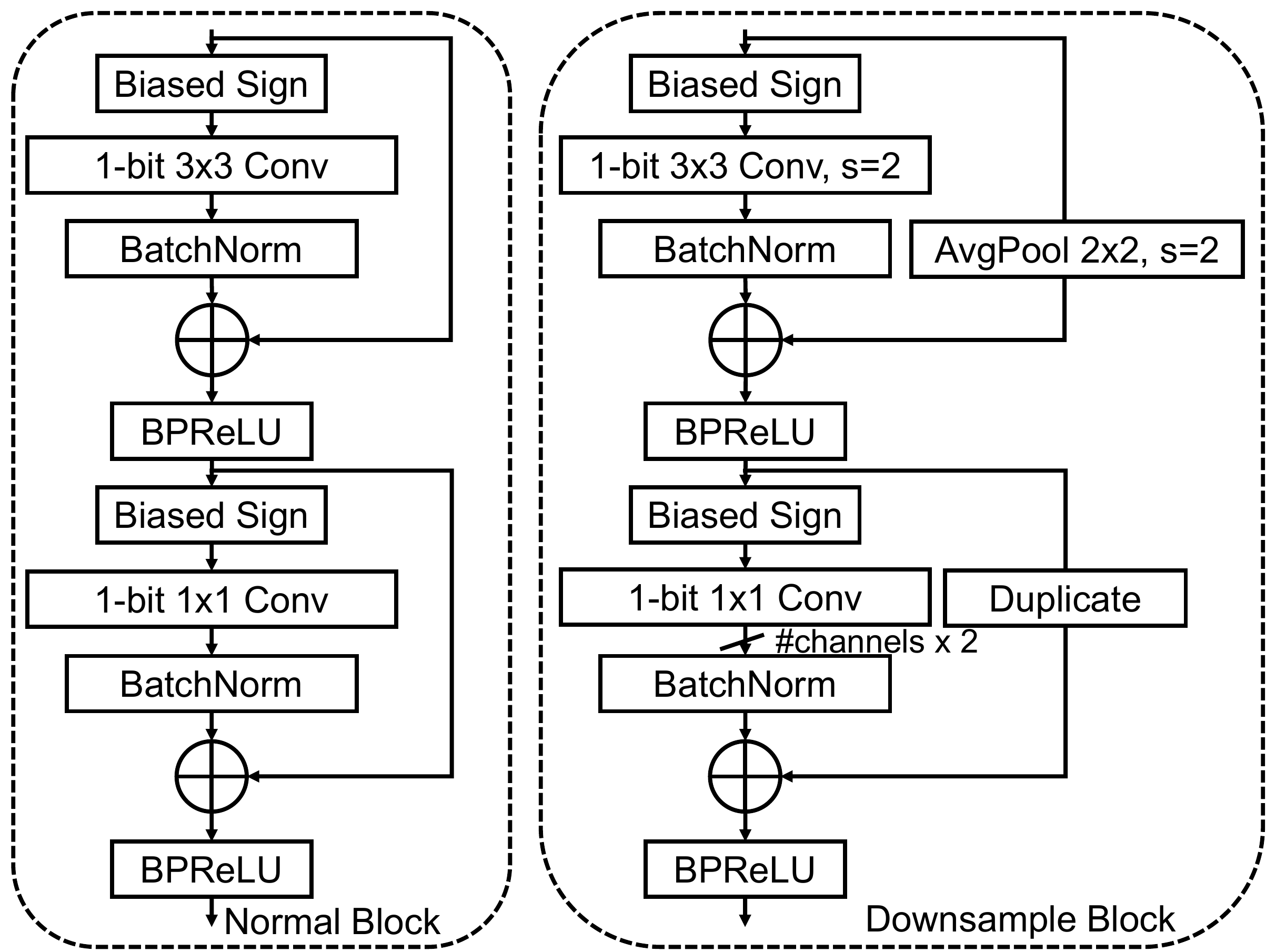}
  \caption{ReActNet building blocks.}
  \label{fig:reactnet-building-block}
  \vspace{-10pt}
\end{figure}
\begin{figure}[ht]
  \centering
  \includegraphics[width=\linewidth]{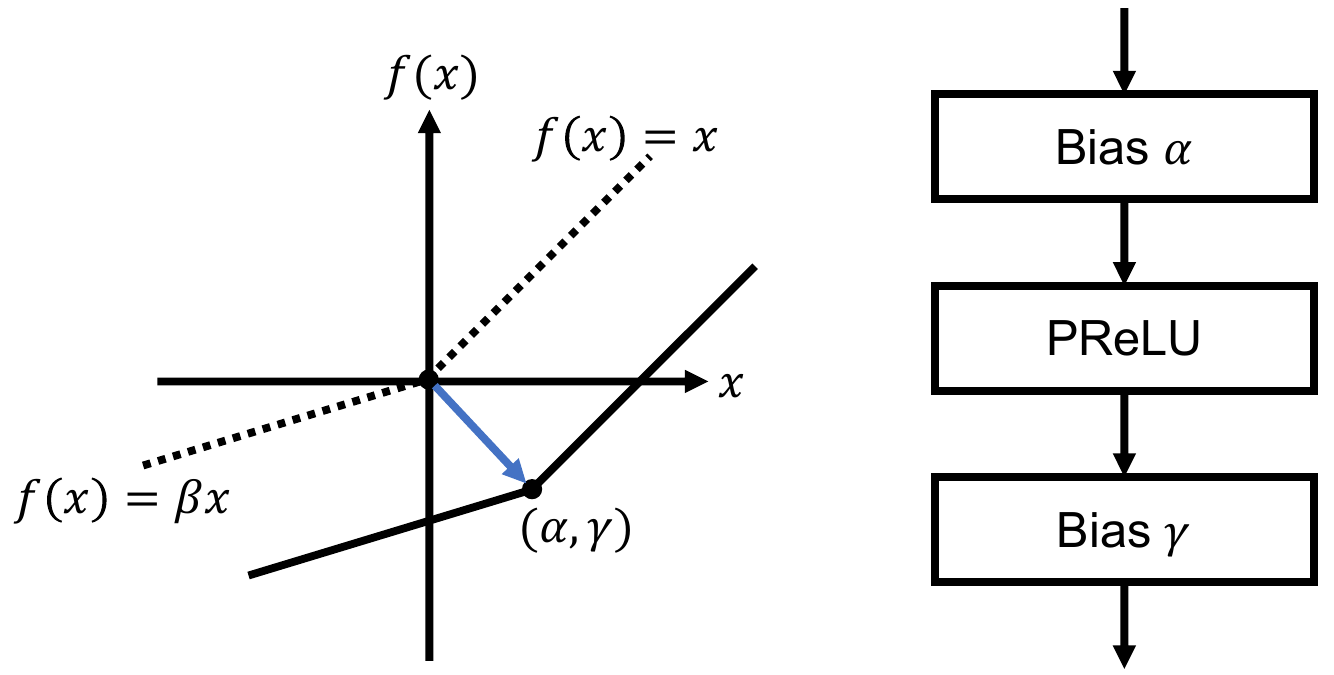}
  \caption{Biased parametric ReLU (BPReLU).}
  \label{fig:biased-prelu}
\end{figure}

Most recently, a more accurate BNN model named ReActNet~\cite{liu2020reactnet}, is proposed to mitigate the accuracy gap between the binarized model and its full-precision counterpart. The building blocks of ReActNet is shown in Figure~\ref{fig:reactnet-building-block}. ReActNet is based on MobileNet~\cite{howard2017mobilenets} architecture. It achieves a top-1 accuracy of 69.4\% on ImageNet dataset using 4.82G BMACs with a model size of 4.6 MB. 

The key feature of ReActNet is a biased PReLU (BPReLU) activation function that shifts and reshapes the feature maps between two convolutional layers~\cite{he2015prelu}. This substantially improves the model accuracy. As shown in Figure~\ref{fig:biased-prelu}, BPReLU translates the PReLU function to a new origin point $(\alpha, \gamma)$. It is implemented as a PReLU function sandwiched by two learnable channelwise biases $\alpha$ and $\gamma$. Based on the same idea, ReActNet introduces a learnable bias to the sign function to learn the binarization threshold through optimization.
Similar to Bi-RealNet~\cite{Liu_2018_ECCV}, ReActNet also adds a full-precision shortcut connection to each convolutional layer in the model. In the downsample layer, the average pooling layer and the channel duplication ensure the shortcut matches the spatial and channel dimensions of the residual. ReActNet uses full instead of depthwise convolutional layers since they increase the capacity of the binarized model. The ReLU activation functions in the original MobileNet are all removed. The limitation of ReActNet is that the input layer is floating-point. Moreover, its accuracy remains low compared to compact networks such as MobileNetV2 which has 72\% top-1 accuracy and 3.4 million parameters.
\section{The Fractional BNN Model}
\label{sec:fracnet-model}
In this section, we will present our fractional BNN model. We first describe how we improve the building block of ReActNet to achieve a higher accuracy. 
We then propose a novel method of binarizing the input layer with minimal accuracy degradation. Finally, we introduce the fractional convolutional layer to further improve model accuracy. 
{\Name} preserves the key hardware benefits of conventional BNNs. Meanwhile, it achieves a top-1 accuracy of 71.8\% on ImageNet, which rivals that of 8-bit MobileNetV2-level with a slightly larger model size.

\subsection{New Building Blocks}
\label{sec:building-block}


\begin{figure}[ht]
  \centering
  \includegraphics[width=\linewidth]{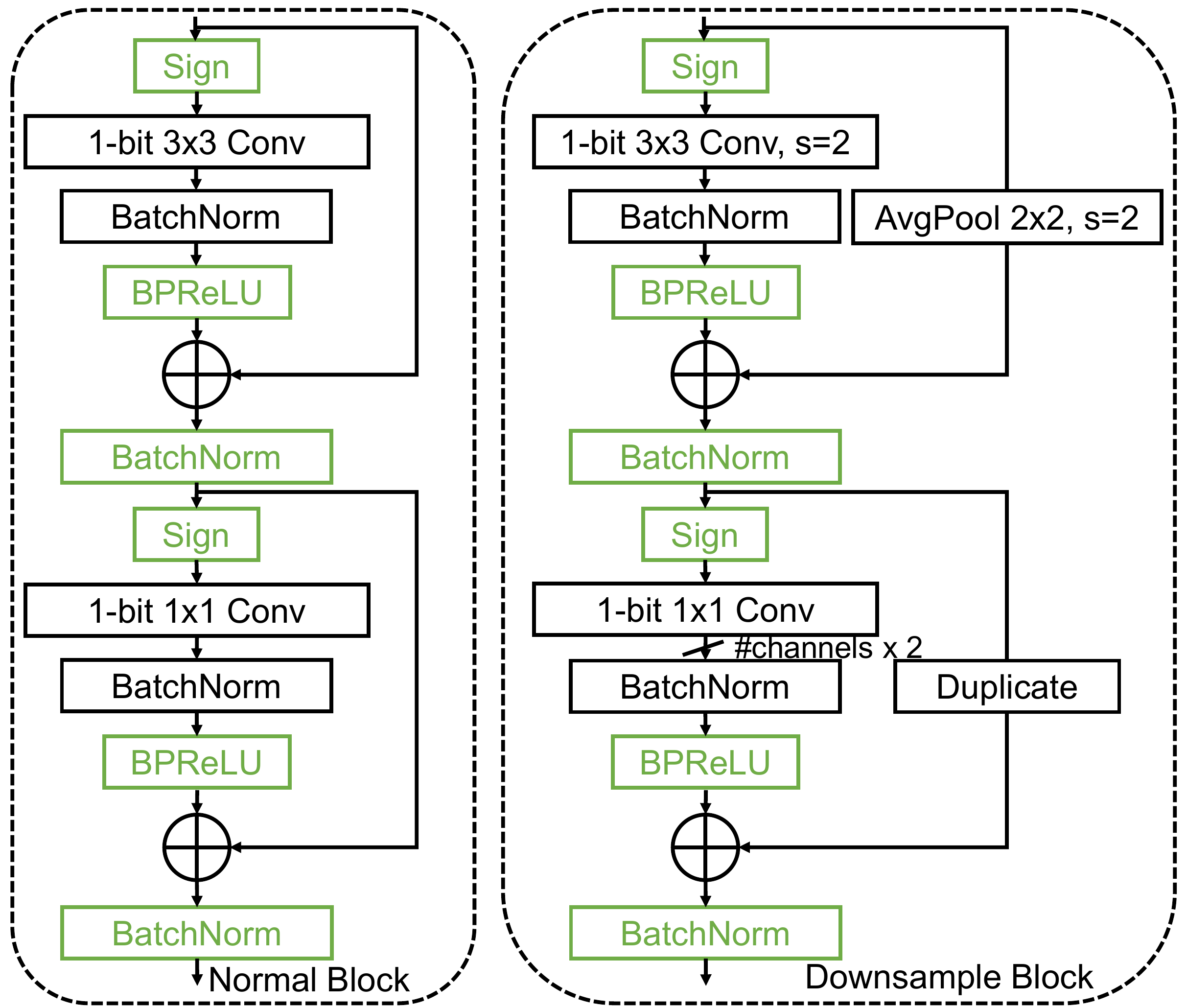}
  \caption{Our model building blocks -- \textnormal{Differences from ReActNet are highlighted}. 
  }
  \label{fig:building-block}
  \vspace{-10pt}
\end{figure}

The building block used in our baseline BNN model is illustrated in Figure~\ref{fig:building-block}.
Different from ReActNet, we move the BPReLUs before the shortcut connections, given that previous works have pointed out that  activation functions residing before the shortcuts tend to perform better~\cite{Martinez2020realtobinary, bulat2019improved}. We also add a BatchNorm layer~\cite{ioffe2015batchnorm} after each shortcut connection. The affine transformation of the BatchNorm layer serves as learning a new distribution for both branches in the next block such that the number of positive and negative values in the activations are more balanced. This shift in distribution is shown to be crucial for feature learning in binary convolutional layers in ReActNet. We generalize it to the shortcut connections as well. Since BPReLU and BatchNorm layers only contain channelwise parameters, their impact on the total number of model parameters is negligible. We further find that a learnable threshold is unnecessary for the sign function after adding the BatchNorm layer since it is already included in the bias term. 

\begin{table}[ht]
  \caption{Comparison with SOTA ResNet-20 BNN on CIFAR-10 dataset -- \textnormal{The first and last layers are in floating-point}.}
  \label{tab:cifar10-bnn}
  \vspace{-5pt}
  \begin{tabular}{c|cccccc}
    \toprule
    Model & Method & Precision (W/A) & Acc. \% \\
    \midrule
          & DoReFa~\cite{zhou2016dorefa} & 1/1 & 79.3 \\
          & DSQ~\cite{gong2019dsq} & 1/1 & 84.1 \\
    ResNet-20 & IR-Net~\cite{qin2019irnet} & 1/1 & 86.5 \\
          & ReActNet~\cite{liu2020reactnet} & 1/1 & 85.8 \\
          & Ours & 1/1 & \textbf{87.2} \\
  \bottomrule
\end{tabular}
\vspace{-5pt}
\end{table}

On the CIFAR-10 dataset, we observe that the ResNet-20 BNN model with our proposed building block outperforms other state-of-the-art binarized ResNet-20 variants.
We modify the popular ResNet-20 model using the top half of the proposed block in Figure~\ref{fig:building-block} that contains a 3$\times$3 convolutional layer. Table~\ref{tab:cifar10-bnn} shows the comparison against other methods. DoReFa~\cite{zhou2016dorefa}, DSQ~\cite{gong2019dsq}, and IR-Net~\cite{qin2019irnet} explore different backward approximations for the Sign function. 
As the result shows, the binarized ResNet-20 model constructed with our proposed building blocks achieves the highest accuracy (87.2\%). The nontrivial 
accuracy improvement over the ReActNet structure is gained by moving the activation function right after the convolutional layer and shifting the distribution of both the convolutional branch and the shortcut. 


\begin{table}[ht]
  \caption{Accuracy on ImageNet of the corresponding real-valued models before binarization.}
  \label{tab:building-block-acc}
  \vspace{-5pt}
  \begin{tabular}{cccc}
    \toprule
    Model & Params  & MACs   & Acc. \% \\
          & ($\times 10^6$) & ($\times 10^9$) & \\
    \midrule
    Full Conv MobileNet~\cite{howard2017mobilenets}  & 29.3 & 4.8 & 71.7 \\
    Real-Valued ReActNet~\cite{liu2020reactnet} & 29.3 & 4.8 & 72.4 \\
    Ours  & 29.3 & 4.8 & \textbf{75.6} \\
  \bottomrule
\end{tabular}
\vspace{-5pt}
\end{table}

For ImageNet dataset, we replace the depthwise separable convolutional layer in the MobileNet architecture with our proposed building block shown in Figure~\ref{fig:building-block}. 

To estimate the potential of the binarized model, we first synchronize the forward and backward behaviors of the sign function to be the straight-through estimator, and train the corresponding real-valued model on ImageNet. The results are in Table~\ref{tab:building-block-acc}. All three models have the same number of parameters and MACs. We observe that our approach is 3.2\% more accurate than the real-valued ReActNet. Compared with the base full convolution MobileNet, the accuracy increment is 3.9\%, even higher than that with ReActNet. The dramatic improvement of our model on ImageNet indicates the efficacy of the pre-shortcut BPReLU activation function and the balance of both branches in a convolutional block. 

\subsection{Binary Input Layer}
\label{sec:binary-input-layer}


\begin{figure}[ht]
    \centering
    \begin{subfigure}[b]{0.45\textwidth}
        \centering
        \includegraphics[width=\textwidth]{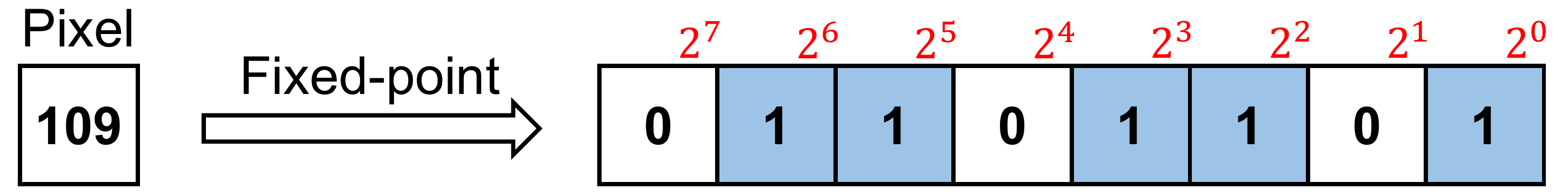}
        \caption{Fixed-point encoding -- \textnormal{Representing a fixed-point number as an 8-dimensional binary vector.}}
        \label{fig:fixed-point-encoding}
    \end{subfigure}
    \hfill
    \begin{subfigure}[b]{0.45\textwidth}
        \centering
        \includegraphics[width=\textwidth]{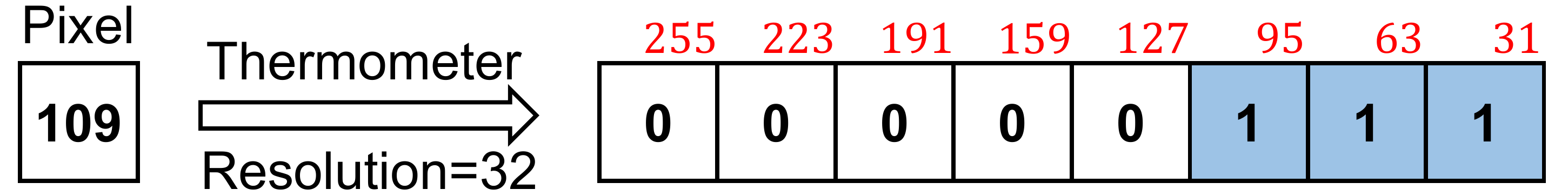}
        \caption{Thermometer encoding -- \textnormal{The ratio of height of '1's in an encoded binary vector reflects the magnitude of a pixel.}}
        \label{fig:thermometer-encoding}
    \end{subfigure}
    \vspace{-10pt}
    \caption{Encoding input images.}
    \label{fig:encode-input-image}
    \vspace{-10pt}
\end{figure}


A binary input layer can reduce the resource consumption of an FPGA accelerator, since a separate floating- or fixed-point convolution engine is no longer required. 
The challenge of binarizing both weights and activations in the input layer is the lack of input channels. 
Anderson et al.~\cite{anderson2018hdgeometry} show that high-dimensional binary vectors can approximately preserve the dot products in the continuous space. Given the 3$\times$3 kernels, the dimensionality of dot products in the input layer depends on the number of input channels. Three RGB channels are insufficient for binarization. It is therefore necessary to split the images into more channels.

Directly using the fixed-point representation of a pixel incurs a  large loss.
As shown in Figure~\ref{fig:fixed-point-encoding}, a natural way of enriching the channels is to treat each pixel as an 8-dimensional binary vector since pixels are 8-bit fixed-point numbers.
Nevertheless, each binary digit has its own associated weight, labeled at the top right corner of each bit. By the time a pixel is converted to a binary vector, the weight information is lost. The magnitude of each bit becomes the same. One may argue that neural network can learn the weights of the binary digits to make an equivalent transformation. However, this is a very challenging task for the BNN training. 

In this work, we propose to use thermometer encoding to transform a pixel to a thermometer vector.
Previous work has used thermometer encoding to resist adversarial attacks to neural networks~\cite{buckman2018thermometer}. There is also a study~\cite{guo2018fbna} that binarizes the input images but the dimension of the encoded vector must be a power of two. Here we use thermometer encoding to binarize the input layer in an end-to-end trainable BNN, and our method supports a flexible vector length.
Given a pixel intensity $p$, $i \in \begin{Bmatrix} 1, \dots, L \end{Bmatrix}$ is the index of its thermometer vector $TV \in {\begin{Bmatrix} 0,1 \end{Bmatrix}}^{L}$, then $TV$ is defined as 
\begin{displaymath}
    TV_{i} = \left\{\begin{matrix}
    0 & 1 \le i \le L-p \\ 
    1 & L-p < i \le L
    \end{matrix}\right.
\end{displaymath}
Namely, the number of $1$s in $TV$ is exactly equal to $p$. The integer $L$ is the dimensionality of $TV$. In this case $L=255$. An input image with RGB channels is now converted to 765 (255*3) binary channels. The dimensionality of the dot products hence increases, and there is no associated weights on each bit.

To provide a flexibility on the dimensionality of $TV$, we further introduce a hyperparameter, resolution $R$. As depicted in Figure~\ref{fig:thermometer-encoding} where $R=32$, each `1' in the encoded thermometer vector represents an intensity of 32. An intensity less than 16 (i.e., $0.5*R$) will be rounded to `0'. Therefore, $p=109$ is converted to a binary vector with three `1's. Formally, the new thermometer vector $\tilde{TV}$ is defined as:
\begin{displaymath}
    \tilde{TV}_{i} = \left\{\begin{matrix}
    0 & 1 \le i \le {\left\lceil{\frac{255}{R}}\right\rceil} - {\left\lfloor{\frac{p}{R}}\right\rceil} \\ 
    1 & {\left\lceil{\frac{255}{R}}\right\rceil} - {\left\lfloor{\frac{p}{R}}\right\rceil} < i \le {\left\lceil{\frac{255}{R}}\right\rceil}
    \end{matrix}\right.
\end{displaymath}
where $\left\lfloor{}\right\rceil$ is the round operation, and $L={\left\lceil{\frac{255}{R}}\right\rceil}$.
Throughout the experiments in this work, we select $R=8$. Hence, each input channel is expanded to 32 binary channels.

Finally, we transform the thermometer encoded input images to the \{-1, +1\} bipolar representation to keep consistent with other activations in the network. Namely, we replace all 0s with $-1$s in the thermometer vectors. The weights in the input layer are binarized using the regular sign function.

\begin{table}[ht]
  \caption{Results of binarizing the input layer using thermometer encoding on CIFAR-10 -- \textnormal{ResNet-20 BNN has 0.27 million parameters and 40.9 million BMACs.}}  
  \label{tab:binary-input-layer-results}
  \vspace{-5pt}
  \begin{tabular}{c|cccccc}
    \toprule
    Model  & Method &  Acc. (\%) & $\Delta$ (\%) \\
    \midrule
               & Base  &  87.2 & 0.0 \\
    ResNet-20  & DBID~\cite{drichen2018binary}   &  78.9 & -8.3 \\
       BNN     & BIL (K=256)~\cite{drichen2018binary}   &  83.7 & -3.5 \\
               & Thermometer (R=8) & 87.2 & \textbf{0.0} \\
  \bottomrule
\end{tabular}
\vspace{-5pt}
\end{table}

The evaluation of binarizing the input layer is in Table~\ref{tab:binary-input-layer-results}. Prior works~\cite{drichen2018binary} attempted direct unpacking of the 8-bit fixed-point input data, dubbed as \textit{DBID}, and adding an additional binary pointwise convolutional layer between the unpacked input data and the first layer to increase the number of channels, dubbed as \textit{BIL}. We implement and compare our proposed method against these techniques on the ResNet-20 BNN introduced in Section~\ref{sec:building-block}.


Our binarized ResNet-20 model has an accuracy of 87.2\%. On such a lightweight model, directly unpacking the 8-bit fixed-point data and binarizing the input layer leads to an 8.3\% accuracy degradation. This shows that the hidden associated weight information is critical to feature learning. On the basis of that, adding an extra pointwise convolutional layer expands the number of channels from 24 to 256 (K=256), thus increasing the model capacity. This does not address the fundamental problem of losing the associated weights, and still results in a 3.5\% degradation. Our proposed method of using thermometer encoding works very well in this case. It preserves the model accuracy after binarizing the first layer.
In the meantime, it has 2.7$\times$ less BMACs in the input layer compared to BIL since our approach only requires 96 channels. 
Hence a binarized input layer with thermometer encoding can enjoy a reduced latency and FPGA resources without sacrificing the accuracy.

\begin{figure}[ht]
  \centering
  \includegraphics[width=\linewidth]{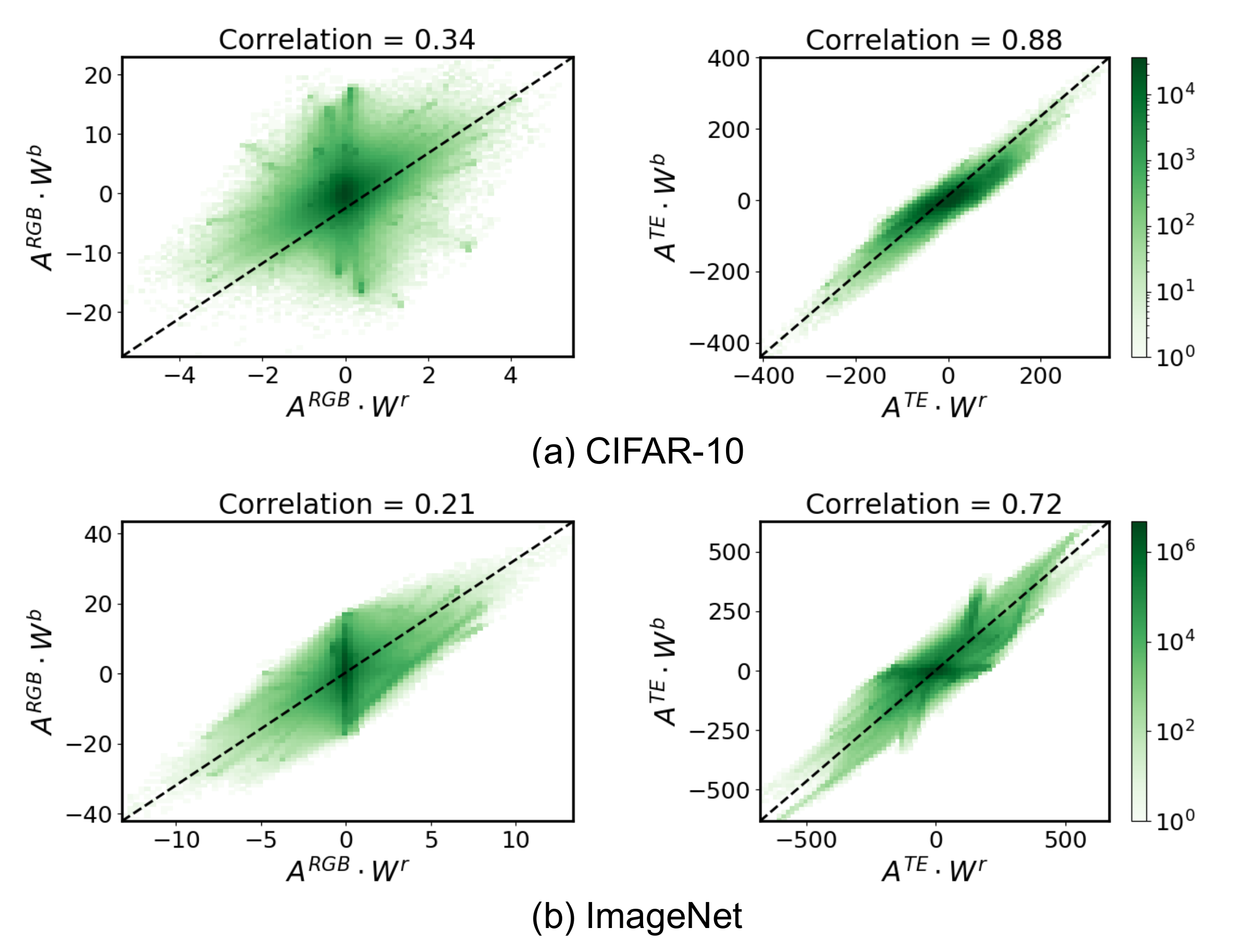}
  \caption{Pearson correlation between dot products before and after binarizing the weights -- \normalfont{Dot products in the left column use the RGB images $A^{RGB}$. Dot products in the right column use the thermometer encoded binary inputs $A^{TE}$.}}
  \label{fig:correlation}
\end{figure}

To understand how an increment in input channels helps with the binarization, we analyze the correlation of the dot products before and after binarization in the input layer. 
In Figure~\ref{fig:correlation}, for both CIFAR-10 and ImageNet models, we plot the 2D histogram of the dot products of the activations and binarized weights (vertical axis) and the dot products of the activations and floating-point weights (horizontal axis) in the input layer. As shown in the first column, the correlation is weak if the inputs are RGB images and have three channels. 
This is consistent with the observation by Anderson et al.~\cite{anderson2018hdgeometry}. While using a 96-channel thermometer encoding in the second column, we see that the pre- and post-binarization dot products are highly correlated. This means that thermometer encoding preserves the feature similarity after binarizing the input layer, thus achieving minimal accuracy degradation.

\subsection{Fractional Convolution}
\label{sec:fractional-conv}

\begin{figure}[ht]
  \centering
  \includegraphics[width=\linewidth]{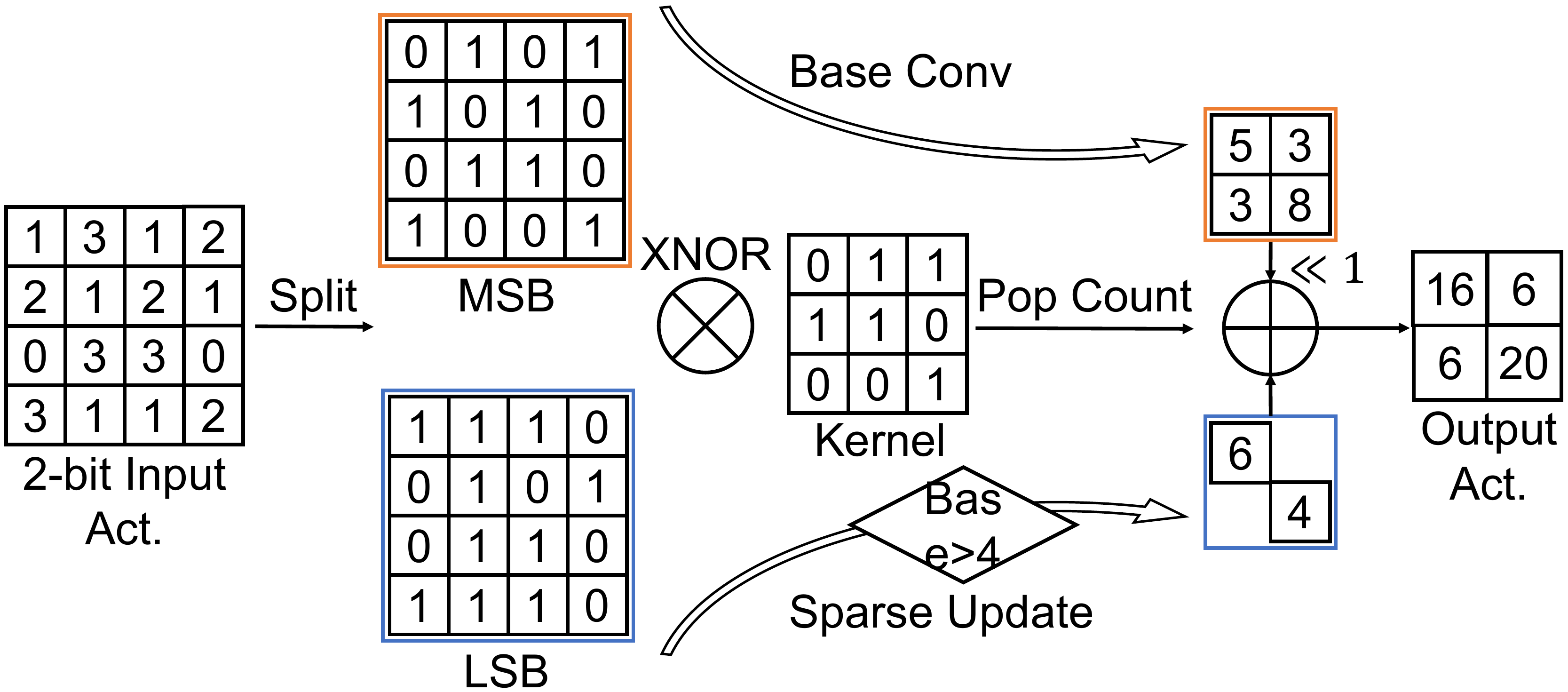}
  \caption{Improving BNN by computing an additional sparse binary convolutional layer.}
  \label{fig:bnn-pg}
  \vspace{-10pt}
\end{figure}

Low-precision quantized networks usually suffer from accuracy degradation compared to their full-precision counterparts~\cite{choi2018pact}. Prior work proposes an end-to-end trainable technique \textit{precision gating} (PG)~\cite{Zhang2020Precision} that dynamically updates important features to high precision to improve the model accuracy. Binarization is an extreme case of quantization. Hence, our fractional convolutional layer (FracConv) adopts PG in the binarized network, except for the input layer and the classifier, to further increase the accuracy. During inference, we dynamically identify important features and compute an additional BMAC to update their popcnt results.

The specific execution flow of FracConv is shown in Figure~\ref{fig:bnn-pg}. Instead of binarizing the activations, we store the 2-bit activations in memory. The weight kernels are 1-bit as usual. Each convolutional layer now consists of a base phase and an update phase. The base phase is a regular binary convolutional layer where the inputs are the 1-bit weights and the most-significant bit (MSB) of the activations. During the computation, we maintain a flag bit for each output feature that indicates whether its popcnt result is larger than a learnable threshold $\Delta$. 
If the flag bit is 1, the output feature is considered to contribute more to the model predictions. It will then be updated by an additional BMAC in the update phase that takes the least-significant bit (LSB) of the activations and the same 1-bit kernels as inputs. Formally, we write the outputs of a FracConv as: 
\begin{displaymath}
    \textbf{O} = \left\{\begin{matrix}
    \textbf{O}_{MSB}<<1 & \textbf{O}_{MSB} \le \Delta \\ 
    (\textbf{O}_{MSB}<<1) + \textbf{O}_{LSB} & \textbf{O}_{MSB} > \Delta
    \end{matrix}\right.
\end{displaymath}
where
\begin{displaymath}
    \textbf{O}_{MSB} = \text{popcnt} ( \text{XNOR} (\textbf{W}^{b}, \textbf{X}_{MSB}^{b} ) )
\end{displaymath}
\begin{displaymath}
    \textbf{O}_{LSB} = \text{popcnt} ( \text{XNOR} (\textbf{W}^{b}, \textbf{X}_{LSB}^{b} ) )
\end{displaymath}
and $<<$ is the left shift operation. The learnable threshold $\Delta$ is channelwise, and can be viewed as scoring the importance of an output feature based on its popcnt result in the base phase.

Although FracConv incurs extra computation, it still enjoys the benefit of the efficient kernels in BNNs. All of the MACs in the convolutional layers in our model are binary.
Moreover, the update phase in FracConv is sparse, thus additional compute effort is not significant. While the activations are  quantized to two bits, the memory footprint of {\Name} is similar to a regular BNN since the weights remain binary. Also note that the popcnt accumulations produce multi-bit integers regardless. Hence the increased size of the activation/feature buffer in the update phase is small compared to the weight storage.  

\begin{table*}
  \caption{Comparison of {\Name} with other efficient models on ImageNet -- \textnormal{The effective bitwidth of activations in {\Name} is 1.44 bits since the average sparsity in fractional convolution is 56\%. IMAC denotes integer MACs. FPMAC denotes floating-point MACs.}}
  \label{tab:fracnn-results}
  \vspace{-5pt}
  \begin{tabular}{cccccccc}
    \toprule
    Network & Precision  & Model Size  & IMAC & BMAC & FPMAC & Top-1 & Top-5 \\
     & (W/A) & (MB) & ($\times 10^{8}$) & ($\times 10^{9}$) & ($\times 10^{8}$) &  (\%) &  (\%) \\
    \midrule
    Bi-RealNet-34~\cite{Liu_2018_ECCV} & 1/1 & 3.18 & 0 & 3.53 & 1.39 & 62.2 & 83.9 \\
    MeliusNet-29~\cite{bethge2020meliusnet} & 1/1 & 5.10 & 0 & 5.47 & 1.29 & 65.8 & 86.2 \\
    MeliusNet-42~\cite{bethge2020meliusnet} & 1/1 & 10.1 & 0 & 9.69 & 1.74 & 69.2 & 88.3 \\
    Real-to-Binary Net~\cite{Martinez2020realtobinary} & 1/1 & 1.92 & 0 & 1.68 & 1.56 & 65.4 & 86.2 \\
    ReActNet-A~\cite{liu2020reactnet} & 1/1 & 4.56 & 0 & 4.82 & 0.12 & 69.4 & - \\
    BNN Ensemble~\cite{zhu2019ensemblebnn} & 1/1 & 11.52 & 0 & 10.10 & 8.34 & 61.0 & - \\
    FP-BNN~\cite{liang2018fpbnn} & 1/1 & 11.4 & 1.10 & 1.03 & 0 & 42.9 & 66.8 \\
    JPEGCompress~\cite{nakahara2020jpegcompression} & 1/8 & 0.72 & 6.21 & 0 & 0 & 70.8 & 90.1 \\
    Synetgy~\cite{yang2019synetgy} & 4/4 & 2.16 & 3.30 & 0 & 0.09 & 68.3 & 88.1 \\
    ResNet-18~\cite{he2016resnet} & 8/8 & 11.69 & 0 & 0 & 18.2 & 69.8 & 89.1 \\
    MobileNet~\cite{howard2017mobilenets} & 8/8 & 4.20 & 0 & 0 & 5.69 & 70.6 & - \\
    MobileNet V2~\cite{Sandler_2018_CVPR} & 8/8 & 3.47 & 0 & 0 & 3.00 & 71.8 & 91.0 \\
    ShuffleNet $1.5 \times$~\cite{zhang2018shufflenet} & 8/8 & 3.40 & 0 & 0 & 2.92 & 71.5 & 90.2 \\
    \cmidrule{1-8}
    \textbf{\Name} & 1/1.4 & 4.56 & 0.01 & 7.30 & 0 & \textbf{71.8} & 90.1 \\
    \bottomrule
  \end{tabular}
  \vspace{-5pt}
\end{table*}

We integrate the binary input layer and fractional convolution into our base BNN model introduced in Section~\ref{sec:building-block} and construct {\Name}. We then evaluate {\Name} on the ImageNet dataset. The results are in Table~\ref{tab:fracnn-results}. We measure that the sparsity in the update binary convolutional layer is 56\%. Hence, the equivalent precision of activations in {\Name} is calculated by $1b + (1-56\%)b \approx 1.4b$. The baselines include state-of-the-art BNNs, low-precision networks from FPGA accelerator designs, as well as some popular full-precision compact models. To estimate the size of the full-precision models, we  assume that they use 8-bit weights and activations since prior studies have shown that 8-bit quantization usually incurs a small accuracy loss \cite{choi2018pact, lin2016fixpoint, zhao2019ocs}. 

From the ImageNet results we have several key observations: 

\textbf{The accuracy of {\Name} significantly outperforms prior BNN and low-precision FPGA accelerators.}
{\Name} is 28.9\% higher in top-1 accuracy than FP-BNN \cite{liang2018fpbnn}, an FPGA BNN accelerator which implements the binarized AlexNet.
Among low-precision networks, DiracDeltaNet constructed in Synetgy~\cite{yang2019synetgy} is based on ShuffleNetV2~\cite{ma2018shufflenetv2}. It replaces the compute-intensive 3$\times$3 convolutional layer by a shift and a pointwise convolutional layer. This results in a higher efficiency for FPGA implementation, but at the expense of reduced model capacity. The JPEGCompress~\cite{nakahara2020jpegcompression} uses a model that is very similar to Synetgy, but is deeper. On the ImageNet dataset, the top-1 accuracy of {\Name} is 3.5\% and 1\% higher than Synetgy and JPEGCompress, respectively. 

\textbf{The accuracy of {\Name} surpasses SoTA BNNs by a large margin.} At a similar model size, {\Name} outperforms MeliusNet-29~\cite{bethge2020meliusnet} and ReActNet-A~\cite{liu2020reactnet} by 6\% and 2.4\% in top-1 accuracy, respectively. Even with a 2.2$\times$ smaller model size, {\Name} is 2.6\% more accurate than MeliusNet-42. Compared to BNN Ensemble~\cite{zhu2019ensemblebnn} that aggregates six binarized ResNet-18 models, {\Name} is 2.5$\times$ smaller in size but 10.8\% higher in accuracy.

\textbf{{\Name} achieves MobileNetV2 level accuracy.} We compare {\Name} with popular full-precision compact network architectures, and observe that {\Name} achieves the same accuracy level as MobileNetV2. While its convolutional layers are computed in pure BMACs, {\Name} can still reach and even surpass the accuracy of compact CNNs such as MobileNet~\cite{howard2017mobilenets} and ShuffleNet 1.5$\times$~\cite{zhang2018shufflenet}. It is also worth noting that {\Name} is more accurate than ResNet-18 (+2\%) with a 2.6$\times$ smaller model size.

\textbf{{\Name} has the lowest number of floating-point MACs.} With the help of the binary input layer and the fractional convolutional layers, the dominant arithmetic operations in {\Name} are BMACs. Only the classifier is computed in integer MACs, and there are no floating-point MACs. Other models in the baselines have floating-point or 8-bit input layers. Though {\Name} has a considerable number of BMACs, they can be massively parallelized on FPGAs.

\begin{table}[ht]
  \caption{Comparison of {\Name} with other BNNs on CIFAR-10 -- \textnormal{The effective precision of activations in {\Name} is 1.4 bits since the average sparsity in fractional convolution is 60\%.}}
  \label{tab:fracnet-cifar10-results}
  \vspace{-5pt}
  \begin{tabular}{ccccc}
    \toprule
    Model   & Model Size & BMAC              & IMAC              & Top-1 \\
            & (MB)       & ($\times 10^{6}$) & ($\times 10^{6}$) & (\%)  \\
    \midrule
    IR-Net~\cite{qin2019irnet}   & 0.03 & 40.5 & 0.4 & 86.5 \\
    FP-BNN~\cite{liang2018fpbnn} & 1.67 & 58.1 & 3.6 & 86.3 \\
    BNN~\cite{zhao2017bnnacc}    & 1.67 & 58.2 & 3.5 & 88.9 \\
    \cmidrule{1-5}
    \textbf{{\Name}} & \textbf{0.03} & 71.5 & 0 & \textbf{89.1} \\
    \bottomrule
  \end{tabular}
  \vspace{-5pt}
\end{table}

We also show the CIFAR-10 results of {\Name} in Table~\ref{tab:fracnet-cifar10-results}. Compared to previous FPGA BNN accelerators~\cite{liang2018fpbnn, zhao2017bnnacc}, {\Name} achieves the highest accuracy, meanwhile with 50$\times$ reduction in model size. This enables fully unrolling the network on an embedded FPGA. With the same model size, {\Name} is also 2.6\% more accurate than IR-Net~\cite{qin2019irnet}, the state-of-the-art ResNet-20 BNN variant.
\section{{\Name} Accelerator Design}

Our {\Name} consists of the following types of operations:
\begin{itemize}
    \item 3$\times$3 fractional convolution (stride 1 and 2)
    \item 1$\times$1 fractional convolution
    \item 3$\times$3 binary convolution (the input layer)
    \item Average pooling
    \item Linear classifier (matrix multiplication)
    \item Batch normalization and BPReLU activation function
    \item Residual connection and concatenation
\end{itemize}

We have designed and implemented an efficient accelerator that supports these operations on an embedded FPGA.
In the following, we will describe the hardware engines in detail.

\subsection{3$\times$3 and 1$\times$1 Fractional Convolution Engine}
\label{sec:3x3frac-conv}
One of the key contributions in our network architecture is the fractional convolutional layer introduced in Section~\ref{sec:fractional-conv}. The fractional convolution scheme improves the accuracy from a single binary convolution with a small resource overhead. 

\begin{figure}[ht]
  \centering
  \includegraphics[width=\linewidth]{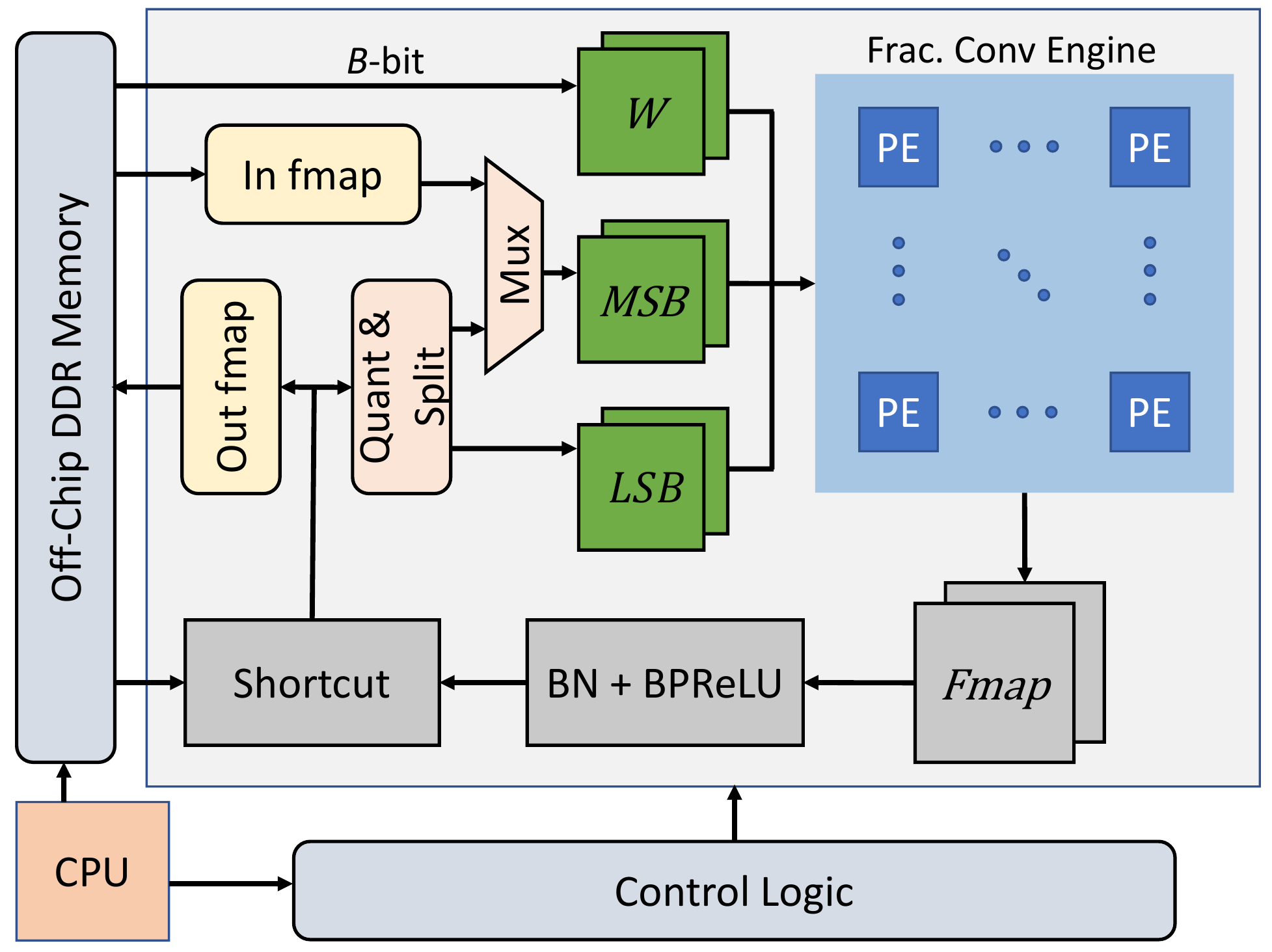}
  \caption{{\Name} accelerator architecture.}
  \label{fig:pgpng}
\end{figure}

Figure \ref{fig:pgpng} illustrates the overall architecture of the accelerator. 
The entire life cycle of a fractional convolution starts with fetching the 2-bit feature maps from the on-chip Block RAM and the 1-bit weights from the off-chip DDR memory. We manage to store the low-precision feature maps on the Block RAM for faster loading. 

After obtaining the feature maps, we first split each 2-bit feature into MSB and LSB. We then pack bits along the channel dimension into $B$-bit arbitrary precision integers for concurrent access. As we acquire weights from the DDR, we pack them into $B$-bit vectors to align the precision with the feature maps. In order to balance between parallelism and resource utilization, we select $B=64$ for the CIFAR-10 design and $B=32$ for the ImageNet design. The packed weights and feature maps are fed to the convolution engine. Meanwhile, auxiliary parameters, including thresholds, and weights in the BatchNorm and activation functions are fetched from the DDR. We also load the fixed-point shortcuts from the DDR to prepare for the residual connection summation.

The pseudo code of the fractional convolution engine is described in Algorithm \ref{alg:fracconv}. As we introduced in Section~\ref{sec:fractional-conv}, the fractional convolution consists of a base phase and an update phase. 

In the base phase (line 5 to 8), the MSB feature maps are convolved with the weights. The computation is the same as a binary convolution. We first compute bitwise XNOR on the $B$-bit features and weights, and then perform a popcnt on the outcome as the final result. We find that the straight-forward implementation for popcnt of directly adding each bit is sufficiently resource efficient. The combined XNOR and popcnt operations are able to be finished in one cycle.
As completing the base path, the output feature maps of the MSB convolution are stored in an on-chip buffer, and we proceed to the update phase.

In the update phase (line 9 to 18), a conditional binary convolution is executed based on the per channel gating thresholds. As shown in the algorithm, an LSB output feature will be computed only when its corresponding MSB output feature exceeds the threshold. The fractional convolution promotes to a binary convolution if the sparsity is $100\%$; on the other hand, it degenerates to a 2-bit convolution if the sparsity is $0\%$. Otherwise it is in between. If the sparsity is low enough, it then becomes more straightforward to directly parallelize the MSB and LSB convolutions to a 2-bit convolution at the cost of more hardware resources.


The binary convolutions are pipelined on the spatial dimension and parallelized on the channel dimension. Within the resource constraint, we fully unroll the computation in a 3$\times$3 window to maximize the parallelism. The point-wise fractional convolution is essentially the same as the 3$\times$3 convolution other than the size of the window. Upon finishing the convolution operations, we perform BatchNorm and BPReLU activation on the results and produces the output of a block. Finally, we organize the outputs into a buffer and transfer them to DDR. 


\vspace{-5pt}

\subsection{3$\times$3 Binary Convolution Engine}
\label{sec:3x3binaryconv}
 The pure binary convolutional layer takes place only at the thermometer encoded inputs. We reuse the 3$\times$3 convolution engine in the fractional convolutional layer to handle this occasion. Since the input layer is excluded from fractional activations, we only execute the base phase.
For ImageNet, we pack 32 bits together along the input channel and feed into the convolution engine. It outputs 32 channels as well. Since the input has 96 channels after the thermometer encoding, we iterate three times for each spatial position and accumulate the results for the output of the layer.

\begin{algorithm}
\caption{Fractional Convolution}
\begin{algorithmic}[1]
\Function{FracConv(fmap, weights, threshold, output)}{}
\State $W \gets $ Width
\State $H \gets $ Height
\State ${\text{fmap}}_{\text{msb}} \gets  $ Split(fmap, msb)
\State ${\text{fmap}}_{\text{lsb}} \gets  $ Split(fmap, lsb)
\For {ch$_{\text{out}}$ in $0$, $1$, ... $C_{\text{out}}-1$}
\For {ch$_{\text{in}}$ in $0$, $1$, ... $C_{\text{in}}-1$}
    \State w $\gets$ LoadWeights(ch$_{\text{out}}$, ch$_{\text{in}}$)
    \State output[ch$_{\text{out}}$] $+=$ BinaryConv(${\text{fmap}}_{\text{msb}}[\text{ch}_{\text{in}}]$, w)
\EndFor
\For {ch$_{\text{in}}$ in $0$, $1$, ... $C_{\text{in}}-1$}
    \State w $\gets$ LoadWeights(ch$_{\text{out}}$, ch$_{\text{in}}$)
\For{pixel in $0$, $1$, ..., $H*W-1$}
    \If {output[ch$_{\text{out}}$][pixel] > \text{threshold[ch$_{\text{out}}$]}} 
        \State act $\gets$ LoadWindow$_\text{lsb}$(${\text{fmap}}_{\text{lsb}}$, ch$_{\text{in}}$, H, W)
        \State output[ch$_{\text{out}}$][pixel] $+=$ \\ popcnt(XNOR(act, w))
    \EndIf
    
\EndFor
\EndFor
\EndFor


\EndFunction
\end{algorithmic}
\label{alg:fracconv}
\end{algorithm}

\subsection{Average Pooling, Classifier, and Other Operations}
In addition to the accelerators for convolution operations, which account for the majority of the computations in the network, we also implement hardware accelerators for less frequent operations such as average pooling and matrix multiplications. 
An average pooling layer sums up features in the spatial dimension, and is divided by the number of features summed together.
The challenge of designing this kernel is determining the amount of parallelism and the trade-off between resource and latency. We use an adder tree to serve as a sum engine to compute the sum in one dimension in a pooling tile in one cycle and then pipeline the addition operation for the other dimension to achieve the optimal concurrency. Even though it is possible to maximize the parallelism by accessing all the elements in a tile at once and compute the sum with the fewest number of cycles, it requires partitioning the array in both dimensions. The cost of resource utilization may not be worth the performance gain. Our implementation achieves comparable performance with economic resource consumption. The matrix multiplication unit serves as the linear classifier at the final state of the network. It computes the class dimension in parallel. For further optimization, it can be assigned to CPU and executed in parallel with the accelerator.
This compact accelerator design is the balanced choice between performance and resource usage. Other miscellaneous operations such as BatchNorm, BPReLU activation, and channel concatenation are fused seamlessly with adjacent operations.


\section{Evaluation}
In this section, we first describe the experiment setup, and then present our results on FPGAs.

\subsection{Modeling Training Setup}
We evaluate FracNN on both CIFAR-10~\cite{krizhevsky2009cifar} and ILSVRC12 ImageNet~\cite{deng2009imagenet} classification datasets. We augment the input images using random horizontal flip and random crop. Color jitter is used only for ImageNet. We follow the two-step training strategy as described in Real-to-Binary Net~\cite{Martinez2020realtobinary}. In the first step, the activations are binarized but the weights are floating-point. The weight decay is 1e-5. In the second step, weights are binarized and initialized from the first step. Activations are still binary. The weight decay is zero. To train the {\Name}, the first two steps are the same except that the activations are quantized to 2 bits. We add a third step that initializes the weights from the second step, and applies the fractional convolutional layer. The first and the last layer are excluded. We use cross-entropy loss during the training on CIFAR-10. For ImageNet, we calculate the KL divergence between the softmax output of a teacher model and that of the trained model as the loss function, same as ReActNet~\cite{liu2020reactnet}. In our experiments the teacher model is a pretrained ResNet-50. 

For CIFAR-10, we train the model for 300 epochs in each training step with a batch size of 128. The initial learning rate is 1e-3, and decays linearly to 0 in each epoch. For ImageNet, we train the model for 120 epochs in each step. The batch size is 256. The initial learning rate is 5e-4 and also decays linearly to 0 in each epoch. We use PyTorch~\cite{adam2019pytorch} to specify models and training scripts. All training experiments are completed on NVIDIA RTX 2080Ti GPUs.

\subsection{FPGA Implementation}

\begin{table*}[ht]
  \caption{Hardware performance of {\Name} on ImageNet at batch size of 1.}
  \label{tab:hardware-imagenet}
  \vspace{-5pt}
  \begin{tabular}{c|cccccccc}
    \toprule
    & ReBNet~\cite{ghasemzadeh2018rebnet} & AlexNet~\cite{liang2018fpbnn} & FINN-R~\cite{blott2018finnr} & T-DLA~\cite{chen2019tdla} & MobileNetV2~\cite{wu2019mobilenetv2fpga} & Synetgy~\cite{yang2019synetgy} & JPEGComp~\cite{nakahara2020jpegcompression} & {\Name} \\
    \midrule
    Device & Virtex & Stratix-V & Zynq  & Zynq  & Zynq  & Zynq  & VirtexUS+  & Zynq\\
           & VCU108 &           & ZU3EG & 7Z020 & ZU2EG & ZU3EG & XCVU9P     & ZU3EG\\
    FPS    & 170    & 862.1     & 200.0 & 20.48   & 205.3      & 41.1    & 3321.2 & 48.1\\
    Top-1 (\%)& 41.43 & 42.9    & 50.3  & 65.6    & 68.1       & 68.3    & 70.8   & \textbf{71.8} \\
    Top-5 (\%)& -   & 66.8      & -     & -       & -          & 88.1    & 90.1   & \textbf{90.1} \\
    Bits (W/A)& 1/1 & 1/1       & 1/2   & 2/2   & 8/8        & 4/4     & 1/8    & 1/1.4 \\
    F$_{\max}$ (MHz) & 200 & 150& 220   & 250   & 430        & 250     & 300    & 250 \\
    Power (W)& -    & 26.2      & 10.2  & 2.58    & -          & 5.5     & 75     & 6.1 \\
    LUT  & 537600   & 230918    & 36249 & 37921   & 31198      & 51776   & 274795 & 50656\\
    BRAM & 3456     & 2210      & 432   & 97      & 145 & 159 & 2746 & 201 \\
    DSP & 768       & 384       & -     & 202     & 212 & 360 & 2370 & 224 \\
  \bottomrule
\end{tabular}
\vspace{-5pt}
\end{table*}

\begin{table}[ht]
  \caption{{\Name} performance on CIFAR-10 (batch size 1).}
  \label{tab:hardware-cifarr}
  \vspace{-10pt}
  \begin{tabular}{c|cccc}
    \toprule
     & ReBNet & BNN & FBNA  & {\Name} \\
     & \cite{ghasemzadeh2018rebnet} & \cite{zhao2017bnnacc} & \cite{guo2018fbna}  & \\
    \midrule
    Device           & Zynq  & Zynq   & Zynq    & Zynq \\
                     & ZC702 & 7Z020  & ZC702   & ZU3EG \\
    FPS              & 2000  & 168.4  & 520.8   & \textbf{2806.9} \\
    Top-1 (\%)       & 86.98 & 88.8   & 88.6    & \textbf{89.1} \\
    Bits (W/A)       & 1/1   & 1/1    & 1/1     & 1/1.4 \\
    F$_{\max}$ (MHz) & 200   & 143    & -       & 250  \\
    Power (W)        & -     & 4.7    & 3.3     & 4.1 \\
  \bottomrule
\end{tabular}
\vspace{-10pt}
\end{table}

We evaluate the performance of the {\Name} accelerator architecture on a Xilinx Ultra96 v2 FPGA board. This board uses the Zynq UltraScale+ MPSoC device (ZU3EG), which contains an embedded ARM CPU. The programmable logic fabric has 71k LUTs, 360 DSPs, and 7.6 Mb BRAMs. Ultra96 v2 is supported by PYNQ
, which can import and invoke the accelerator as an overlay in a Python environment. Programmers can feed data and control signals from software via AXI/DMA interfaces to the accelerator. We have implemented accelerators for both CIFAR-10 and ImageNet. The CIFAR-10 accelerator is fully unrolled on the board. Both designs run at a clock frequency of 250MHz. The post-implementation resource utilization is summarized in Table \ref{tab:hardware-both}.

\begin{table}[ht]
  \caption{Resource utilization of the {\Name} accelerator.}
  \vspace{-10pt}
  \label{tab:hardware-both}
  \begin{tabular}{c|cccc}
    \toprule
     & DSPs & BRAM & LUTs \\
    \midrule
    CIFAR-10     & 126 (35\%) & 212 (98.1\%)   & 51444 (72.9\%) \\
    ImageNet    & 224 (62.2\%)  & 201 (93.0\%) & 50656 (71.8\%) \\
  \bottomrule
\end{tabular}
  \vspace{-10pt}
\end{table}

Table \ref{tab:hardware-imagenet} compares our ImageNet accelerators against other previous works. Notably, our ImageNet model provides a significant advantage in the model accuracy.
The top-1 accuracy is the best amongst the other comparable network models that are implemented on FPGAs. 
To map the entire model on the FPGA, it takes $72\%$ LUTs and $93\%$ BRAM usage. Our DSP utilization is allocated mainly for the index calculation, normalization, and activation functions. Due to the limited available resource, we store the intermediate feature maps on the DDR memory upon completing a combination of convolutional, BatchNorm, and BPReLU layers, and fetch them when computing the residual connection. 
We are able to achieve 48.1 frames per second (FPS) as we test our design in the PYNQ environment. 
Our hardware logic runs at 16 ms. We allocate double buffers to overlap part of the communication overhead and compute. 
Our design can perform real-time image classifications, and the attainable frame rate surpasses current state-of-the-art embedded hardware accelerators on the same task. In Table \ref{tab:hardware-imagenet}, the designs that have significantly higher frame rates either target a server-class FPGA such as Intel Stratix V or have a lower accuracy than ours.
It is worth noting that Synetgy~\cite{yang2019synetgy} take a different approach in their FPGA implementation --- the system consists of independent accelerators, which the CPU invokes them as needed. 

Table \ref{tab:hardware-cifarr} compares our results against other works that target CIFAR-10. {\Name} once again achieves the best model accuracy. Compared to the BNN accelerator in \cite{zhao2017bnnacc} and FBNA \cite{guo2018fbna}, our design achieves the highest frame rate with a better accuracy on a comparable embedded FPGA platform. 
Since our CIFAR-10 network model is very compact, we are able to unroll the entire network on the FPGA logic to eliminate unnecessary transactions between the logic blocks and the DDR memory. The only data transmissions are the input image and prediction results. Under this setting, we achieve an FPS of 2806.9 with 72.9\% of LUT utilization. The frame rate is 1.4$\times$ higher than ReBNet~\cite{ghasemzadeh2018rebnet}, which has the highest frame rate among the baselines but with a lower accuracy (by 2.1\%). 


\subsection{Ablation Study}
We further validate the efficacy of the binary input layer and the fractional convolutional layer.

We observe that our binary input layer runs significantly faster than a conventional fixed-point input layer with trivial additional resource consumption.
Table~\ref{tab:abla2} shows the comparison in resource utilization and latency. As described in Section~\ref{sec:3x3binaryconv}, our binary input layer reuses the convolution engine in the fractional convolutional layer. The only resource consumption occurs in loading the image and the weights, therefore remains low. Moreover, the latency of our binary input layer is very low since the convolution engine is highly parallelized when designed for the fractional convolution. In contrast, the conventional input layer with 8-bit inputs and weights requires DSP to compute the the results. The number of DSPs, however, is limited on the target FPGA device. This resource constraint makes it difficult for the 8-bit design to achieve the same parallelism as binary computations, thus resulting in a much higher latency.


\begin{table}[ht]
  \caption{ Comparison between the implementation of our binary input layer and an 8-bit conventional input layer. }
  \vspace{-5pt}
  \label{tab:abla2}
  \setlength{\tabcolsep}{5pt} 
  \begin{tabular}{c|cccc}
    \toprule
    Conv.    & DSPs  & BRAM & LUTs & Latency(ms)\\
    \midrule
    1-bit    & 3 (0.8\%)      & 2.5 (1.2\%)    & 3603 (5.1\%)  &  2.0 \\
    8-bit    & 287 (79.7\%)   & 34 (15.6\%)    & 22509 (31.9\%) &  65.9 \\
  \bottomrule
\end{tabular}
  \vspace{-5pt}
\end{table}


To evaluate the efficiency of the fractional convolution, we implement and compare the logic part of three networks --- each with 1-bit convolutional layers, fractional convolutional layers, and 2-bit convolutional layers, respectively. In the previous sections, we discuss the expectation that a neural network exploiting the fractional convolution should perform slightly worse than a pure binary (1-bit weights and 1-bit activation) model, but have significant improvements over a conventional 2-bit activations and 1-bit weights model. Table~\ref{tab:abla} shows a side-by-side comparison among the three models. Clearly, the implementation results confirm our expectations on the fractional network. We expend some extra resource to ensure the same concurrency as the 1-bit model to match the performance. The model with fractional convolutional layers has a slightly worse latency compared to the 1-bit network, but it is more than 3x better than a conventional 2-bit convolutional network. The only difference in the models is the precision of the convolution accelerator modules. The reason behind such difference is that the conventional 2-bit convolution accelerator requires more resources than either 1-bit or fractional ones. If we hope to ensure the same concurrency, there is no room to fit a much more expensive 2-bit convolution accelerator. Eventually, we have to tune down the concurrency to fit the network on the hardware, resulting in a higher latency but slightly less resource utilization than the 1-bit and fractional convolution models.
\begin{table}[ht]
  \caption{ Comparisons among the 1-bit, fractional, and 2-bit Networks on ImageNet.}
  \vspace{-5pt}
  \label{tab:abla}
  \begin{tabular}{c|cccc}
    \toprule
    Network  & DSPs  & BRAM & LUTs & Latency \\
    Bitwidth &       &      &      & (ms)    \\
    \midrule
    1-bit    & 87 (23.7\%)    & 137 (63.4\%)  & 59488 (84.3\%)  &  15.0 \\
    2-bit    & 85 (23.8\%)    & 169 (78.5\%)  & 53419 (75.7\%)  &  61.3 \\
    Frac.    & 224 (62.2\%)   & 201 (93.0\%)  & 50656 (71.8\%)  &  16.3 \\
  \bottomrule
\end{tabular}
  \vspace{-10pt}
\end{table}

\section{Related Work}

\noindent\textbf{Binary Neural Networks.}
The pioneering works on BNN~\cite{hubara2016bnn, cheng2015ebp} establish the end-to-end training flow for the discrete networks.
Courbariaux et al.~\cite{hubara2016bnn} binarize the weights and activations using the sign function. 
This incurs nearly no loss in accuracy on small datasets such as MNIST~\cite{lecun2010mnist}, SVHN~\cite{netzer2011svhn}, and CIFAR-10~\cite{krizhevsky2009cifar}.
While its preliminary result of binarized AlexNet~\cite{krizhevsky2012alexnet} only achieves 36.1\% top-1 accuracy on the  ImageNet dataset, this work demonstrates the feasibility of BNNs. 
There have been extensive follow-up efforts to improve the accuracy of BNNs. Most of the attempts are along the line of modifying BNN network architectures~\cite{rastegari2016xnornet, Liu_2018_ECCV, bethge2020meliusnet}. 
 There are also works that explore different training strategies of BNNs~\cite{tang2017trainbnn, zhuang2018effective,Martinez2020realtobinary}. 
Recently, PReLU is found to be a better activation function for BNNs~\cite{bulat2019improved}. With an additional shift on the basis of PReLU, ReActNet~\cite{liu2020reactnet} binarizes MobileNet~\cite{howard2017mobilenets} and obtains ResNet-18 level accuracy. 
In addition, \cite{zhu2019ensemblebnn} explores using an ensemble of multiple BNN models to improve the accuracy, albeit at the cost of higher compute complexity. There is also a study that tailors BNNs for FPGAs by constructing LUTNet, an area-efficient LUT-based neural network~\cite{wang2019lutnet}. Unlike the aforementioned research, the proposed approach exploits fractional activations to achieve efficient and accurate quantization. Specifically, we leverage our recent work on precision gating  (PG)~\cite{Zhang2020Precision}, a dynamic dual-precision scheme that updates important features to a high precision at the inference time based on a learnable threshold. We adopt PG in our baseline BNN model motivated by ReActNet, and update a small portion of features to two bits to improve the model accuracy. 

\smallskip
\noindent\textbf{Quantization of Input Layer.}
One visible drawback about current BNNs in terms of hardware design is that the first layer remains full-precision. 
Hirtzlin et al.~\cite{hirtzlin2019stochastic} propose to use stochastic computing for the binarization of the input images. 
This method expands the 3 input color channels from images in CIFAR-10 to more than 1500 binary channels. Consequently, it increases the number of parameters and MACs in the input layer by nearly $16\times$. Dürichen et al.~\cite{drichen2018binary} discuss two other options. The first one is using the 8-bit fixed-point representation of a pixel, named DBID. However, the associated weight of each binary digit is lost when converted to a binary vector. 
In the second option, a pointwise convolutional layer is added between the images and the input layer while using DBID. It does not address the fundamental problem in DBID. 
Unfortunately, when applied to the VGG-8 model on CIFAR-10, these two methods degrade the accuracy by at least 4.6\%. 
Our method is very different from these techniques as we use thermometer encoding to split the pixels into a binary vector. It incurs minimal or even no accuracy degradation.
FBNA~\cite{guo2018fbna} expands each pixel to a sum of a binary vector for a pretrained model; but the dimension of that vector must be a power of two. Our proposed method is different as it can encode a pixel to an arbitrary dimension between 1 and 255, and BNNs with it are still end-to-end trainable.
In MeliusNet~\cite{bethge2020meliusnet}, a grouped stem architecture is proposed to reduce the MACs in the input layer by 40\%. The technique replaces the $7 \times 7$ convolutional layer with three $3 \times 3$ group convolutional layers. Though there is MACs reduction, the input layer is still floating-point. Our proposed method is different as we use a binary input layer.

\smallskip
\noindent\textbf{FPGA-Based CNN Accelerators.}
There have been extensive studies on accelerating low-precision neural networks~\cite{qin2020survey}.
Qiu et al.~\cite{qiu2016vggfpga} first show convolutional layers are computation-bound while fully-connected layers are memory-bound, and propose a dynamic-precision data quantization method to accelerate CNN.
Different hardware architectures are then proposed to address the bottleneck in computation and memory bandwidth~\cite{nakahara2020jpegcompression,zhang2017opencl,wei2017systolic},
and high-level programming frameworks are proposed to help efficient quantization~\cite{naveen2016opencl,zhang2016caffeine,blott2018finnr,gong2020vecq,chen2019tdla}.
However, most of these works target AlexNet and VGG, which are much less efficient than more recent CNN models such as ResNet and MobileNet. 
As another approach, algorithm-hardware co-design method is leveraged to develop hardware-efficient networks with fewer bits in weights and activations~\cite{jiao2017lowbitaxel,yang2019synetgy,wu2019mobilenetv2fpga,guo2017codesign, hao2019iotedge, wang2018hybrid}. 
Our work is very different as all of the convolutional layers in the model are computed in pure BMACs.


There is also an active body of research on accelerating BNN models on FPGAs. 
Zhao et al.~\cite{zhao2017bnnacc} make the first attempt implementing a BNN accelerator on FPGA, which introduces a BitSel module and variable-width length buffers to make the network inference efficient.
FINN~\cite{yaman2017finn, blott2018finnrtrts} provides a framework for fast BNN inference.
ReBNet~\cite{ghasemzadeh2018rebnet} leverages multi-level residual binarization to improve the accuracy.
FBNA~\cite{guo2018fbna} binarizes all the network layers but only targets the CIFAR-10 dataset.
Liang et al.~\cite{liang2018fpbnn} implement the binarized AlexNet on ImageNet. The top-1 accuracy is 42.9\%, which is 13\% lower than its full-precision model.
Our work uses fractional activations to create an accurate model.

\section{Conclusions}

This work proposes {\Name}, which exploits fractional activations to substantially improve the accuracy of BNNs. {\Name} employs a dual-precision activation scheme. Features are computed with up to two bits, using an additional sparse binary convolution. The input layer is also binarized using a novel thermometer encoding. {\Name} preserves the key hardware benefits of conventional BNNs. We design an efficient FPGA-based accelerator for our novel fractional convolution kernel. We also implement the entire optimized network on an embedd FPGA (Xilinx Ultra96 v2). Experiments show that {\Name} achieves a top-1 accuracy comparable to MobileNetV2, surpassing that of the best-known BNN FPGA accelerator and a recently introduced BNN by a large margin. On the embedded FPGA, {\Name} demonstrates the ability of real-time image classification.
\section*{Acknowledgements}
\label{sec-acknowledgements}
This work was supported in part by the Semiconductor Research Corporation (SRC) and DARPA, NSF Award \#2007832, the Xilinx Center of Excellence and Xilinx Adaptive Compute Clusters (XACC) program at the University of Illinois Urbana-Champaign. 
One of the Titan Xp GPUs used for this research was donated by the NVIDIA Corporation. 
We thank Yuwei Hu, Yuan Zhou, Yi-Hsiang Lai, Hanchen Jin, and Ecenur Ustun of the Zhang Research Group at Cornell for their helpful discussions.


\bibliographystyle{plain}
\bibliography{research}


\end{document}